\pdfoutput=1
\PassOptionsToPackage{unicode}{hyperref}
\PassOptionsToPackage{hyphens}{url}
\PassOptionsToPackage{dvipsnames,svgnames,x11names}{xcolor}
\documentclass[
  11pt,
  a4paper,
]{article}
\usepackage{amsmath,amssymb}
\usepackage{iftex}
\ifPDFTeX
  \usepackage[T1]{fontenc}
  \usepackage[utf8]{inputenc}
  \usepackage{textcomp} 
\else 
  \usepackage{unicode-math} 
  \defaultfontfeatures{Scale=MatchLowercase}
  \defaultfontfeatures[\rmfamily]{Ligatures=TeX,Scale=1}
\fi
\ifPDFTeX\else
\fi
\IfFileExists{upquote.sty}{\usepackage{upquote}}{}
\IfFileExists{microtype.sty}{
  \usepackage[]{microtype}
  \UseMicrotypeSet[protrusion]{basicmath} 
}{}
\makeatletter
\@ifundefined{KOMAClassName}{
  \IfFileExists{parskip.sty}{%
    \usepackage{parskip}
  }{
    \setlength{\parindent}{0pt}
    \setlength{\parskip}{6pt plus 2pt minus 1pt}}
}{
  \KOMAoptions{parskip=half}}
\makeatother
\usepackage{xcolor}
\usepackage[margin=2.5cm,top=2.6cm,bottom=2.6cm]{geometry}
\usepackage{graphicx}
\makeatletter
\def\maxwidth{\ifdim\Gin@nat@width>\linewidth\linewidth\else\Gin@nat@width\fi}
\def\maxheight{\ifdim\Gin@nat@height>\textheight\textheight\else\Gin@nat@height\fi}
\makeatother
\setkeys{Gin}{width=\maxwidth,height=\maxheight,keepaspectratio}
\makeatletter
\def\fps@figure{htbp}
\makeatother
\providecommand{\tightlist}{%
  \setlength{\itemsep}{0pt}\setlength{\parskip}{0pt}}
\usepackage[british]{babel}

\def\languageshorthands#1{}

\usepackage{newunicodechar}
\newunicodechar{−}{\textminus}
\newunicodechar{θ}{\ensuremath{\theta}}
\newunicodechar{ω}{\ensuremath{\omega}}
\usepackage{microtype}
\usepackage{setspace}
\usepackage{xcolor}
\definecolor{placeholder}{HTML}{D35400}
\definecolor{linkblue}{HTML}{1F4E79}
\usepackage{xurl}
\usepackage{caption}
\usepackage{titlesec}
\titleformat{\section}{\large\bfseries}{}{0pt}{}
\titlespacing*{\section}{0pt}{16pt}{6pt}
\titleformat{\subsection}{\normalsize\bfseries\itshape}{}{0pt}{}
\titlespacing*{\subsection}{0pt}{10pt}{4pt}
\usepackage{fancyhdr}
\fancypagestyle{plain}{\fancyhf{}\fancyfoot[C]{\small\thepage}}
\fancypagestyle{first}{\fancyhf{}\fancyfoot[C]{\small\thepage}}
\newenvironment{abstractbox}{\par\small\setstretch{1.12}\leftskip=1.2em\rightskip=1.2em}{\par}

\usepackage{etoolbox}
\AtBeginEnvironment{longtable}{\setstretch{1.0}}
\usepackage{longtable,booktabs,array}\usepackage{placeins}

\ifLuaTeX
  \usepackage{selnolig}  
\fi
\IfFileExists{bookmark.sty}{\usepackage{bookmark}}{\usepackage{hyperref}}
\IfFileExists{xurl.sty}{\usepackage{xurl}}{} 
\hypersetup{
  pdftitle={Verification Learning Theory: A Process Model of Checking and Retention in AI-Assisted Judgment},
  pdfauthor={Christoph Trattner},
  pdflang={en-GB},
  colorlinks=true,
  linkcolor={linkblue},
  filecolor={Maroon},
  citecolor={linkblue},
  urlcolor={linkblue},
  pdfcreator={LaTeX via pandoc}}

\author{}
\date{}

\begin{document}

\thispagestyle{first}
\begin{center}
{\LARGE\bfseries Verification Learning Theory: A Process Model of Checking and Retention in AI-Assisted Judgment\par}
\vspace{14pt}
{\large \textbf{Christoph Trattner}\par}
\vspace{6pt}
{\small Department of Information Science and Media Studies, University of Bergen, Norway\par}
\vspace{2pt}
{\small ORCID \href{https://orcid.org/0000-0002-1193-0508}{0000-0002-1193-0508} · Correspondence: christoph.trattner@uib.no\par}
\vspace{8pt}
{\small\itshape Preprint, version 1 (October 2026), not peer reviewed.\par}
\end{center}
\vspace{10pt}
\begin{abstractbox}
\noindent{\bfseries Abstract}\par\smallskip
\noindent Accurate AI assistance can improve immediate judgments while reducing opportunities to practise the operations needed for independent performance. Verification Learning Theory (VLT) formalises this possibility for claim-evaluation practice followed by correct, independent, rule-specific feedback. The model links perceived assistant reliability to evidence checking and links the history of those checks to retained rule use. Its strict form assumes that task demands and perceived reliability influence learning only through checking, with equal learning per genuine check. It therefore predicts that perceived fallibility can improve delayed unaided judgment even without observed assistant errors, and that this advantage disappears when checking is already required and performed. Rule-specific gains and forecasts from checking histories provide further tests. Simulations examine the consequences and parameter dependence of these commitments; they do not constitute empirical validation. Comparisons with depth-of-processing and feedback-encoding accounts identify where predictions diverge and where the accounts remain observationally equivalent. The framework specifies experiments that can test the proposed mechanism and separates retention of practised rules from the broader problem of recognising which rule applies to a new claim.
\par\medskip\sloppy\noindent{\bfseries Keywords:} AI assistance, verification, learning, retention, trust in automation, memory model, misinformation
\end{abstractbox}
\clearpage

\hypertarget{the-learning-problem}{%
\section{1 The learning problem}\label{the-learning-problem}}

Two people can reach the same correct answer with an AI assistant while doing different parts of the thinking themselves. Months later, without the assistant, are both equally able to judge? The question matters because hundreds of millions of people now use AI assistants every week (Chatterji et al., 2025), and help that raises performance while it is present can leave less behind once it is gone. In a randomised experiment in high-school mathematics, students who practised with an unrestricted chatbot did worse on a later exam taken without it than students who practised without one, whereas a version that gave hints instead of answers largely avoided this cost (Bastani et al., 2025). In a preregistered experiment, management consultants given GPT-4 did data-science tasks better while they had it, but on questions answered afterwards without it they knew no more than consultants who had worked with ordinary online resources (Wiles et al., 2026). A preprint reports a similar cost in programming (Shen \& Tamkin, 2026), and an observational study reports one in colonoscopy (Budzyń et al., 2025). Accurate assistance can thus improve immediate judgments while reducing the practice on which independent performance depends.

This article develops that possibility for a task that people already hand to assistants: judging whether a claim is supported by its evidence. An assistant compares a claim, for example that a food prevents a disease, with the evidence and returns a verdict: the evidence supports the claim, contradicts it or is insufficient to judge it. In one month-long study without a comparison group, people's unaided accuracy on new claims fell over four weeks, although dialogues with an AI had improved their judgments while it was present (Rani et al., 2026). The present account concerns a controlled instructional setting: people judge a claim against supplied evidence and then receive a correct, independent explanation of the relevant rule. The formal model predicts retention and use of a specified rule. It does not yet explain how people identify the appropriate rule when unfamiliar claims offer ambiguous or competing cues. Transfer across domains is therefore a proposed test of the account's reach, not an established result of the simulation.

Consider the claim that a sports drink improves the performance of adults who do not exercise regularly, when the only evidence is a trial in trained athletes. The trial does not show what happens in people who do not train, so the warranted response is to withhold judgment. Someone who remembers that `the drink works' has stored an answer. Someone who has learned to ask whether the people studied are those the claim is about has learned a rule, a test of the evidence that works on any topic. In argumentation theory, a warrant is what licenses the step from evidence to claim (Toulmin, 1958), and the rules studied here are tests of whether a warrant holds.

Whether a person applies such a rule depends on what the situation asks for. Suppose the assistant's verdict on the sports-drink claim is that the evidence is insufficient. A \emph{verdict task} asks only for the person's own judgment, and someone who trusts the assistant can complete it by copying the verdict without reading the evidence. A \emph{flip task} offers several changes in how the evidence was obtained, each leaving every finding as it is, and asks which of them would change the response the evidence warrants. Only one does, here a trial run in adults who do not exercise regularly; the others change features that other rules concern without breaking those rules. The verdict does not contain this answer, so the person has to apply the rule to give it. I call a task of this kind \emph{check-requiring}: it cannot be answered correctly without the rule's check, even when the verdict is right. A task that can be answered correctly from a right verdict, such as the verdict task, I call \emph{copyable}.

A \emph{check} is an attempted application of the relevant evidential rule: an examination, before the explanation is read, of whether the evidence meets the rule's condition, here whether the people studied are those the claim is about. It is defined by the attempt and not by its result, so a genuine attempt that reaches the wrong answer is a check, whereas a correct answer copied from the assistant is not. The outcome of interest is \emph{delayed unaided judgment}: accepting, rejecting or withholding judgment on new claims, on topics not practised, as their evidence warrants, months after practice and without the assistant. I call the ability this outcome measures \emph{residual warrant}, the kind of residue that working with a tool can leave in its user (Salomon et al., 1991). It differs in two ways from retained accuracy in using a rule, which asks whether people can still apply a rule they are told applies. People must recognise unprompted that a new claim calls for the rule, and they must accept or reject claims according to the evidence that passes the rule's check, and withhold judgment when none passes.

The assistant's reliability can do the same work as the task. People who have found an assistant to be always right can sensibly copy its verdicts (Lee \& See, 2004; Parasuraman \& Riley, 1997), and people who have seen an automated aid err check its advice more (Bahner et al., 2008). An assistant that is always right may therefore teach people less about judging for themselves than one known to err, even when every verdict is followed by a correct explanation. I call this possibility the \emph{accuracy paradox}. Verification Learning Theory (VLT) states when it should occur and when it should not. Its central restriction is that, with correct feedback held constant, the task and the assistant's perceived reliability affect what people retain only through the checks they prompt, and that a genuine check teaches the same whatever prompted it. Section 2 separates what VLT inherits from what it adds; Section 3 states the process model; Section 4 derives its predictions and sets them against the closest alternative accounts; Section 5 separates what the model's simulation shows by construction from what depends on parameter choices; Section 6 describes a test of the account; and Section 7 discusses its scope.

\hypertarget{from-reliance-to-retained-learning}{%
\section{2 From reliance to retained learning}\label{from-reliance-to-retained-learning}}

VLT inherits three principles from earlier research and adds one claim of its own.

\textbf{Inherited principles.} The first principle comes from research on reliance on automated aids. People rely more on an aid the more reliable they believe it to be and the less they trust their own judgment (Lee \& Moray, 1994; Lee \& See, 2004), and they monitor it less once it has proved reliable (Parasuraman \& Manzey, 2010). People who have experienced an aid's failures check its advice more than people who were only told about them (Bahner et al., 2008), and people revise a stated accuracy in the light of the accuracy they observe (Yin et al., 2019). In the first of three preregistered experiments reported in a preprint, AI advice raised accuracy by about 25 percentage points when it was right and lowered it by about 15 when it was wrong (Shaw \& Nave, 2026). Asking people to judge before they see the AI's answer reduces such overreliance (Buçinca et al., 2021). Explanations that contrast the right answer with a likely misconception have improved later decisions made without the AI (Buçinca et al., 2025), but a recommendation can keep people from learning. In three experiments on judging the nutrients in meals, people learned from an AI's explanations when it gave no recommendation, and from a correct explanation given after their own answer. When they first decided and then saw the AI's recommendation and explanation, they learned no more than from being told whether they were right (Gajos \& Mamykina, 2022). That result marks the boundary of the setting studied here: the explanation came from the AI together with its recommendation, before people learned whether they were right, whereas VLT concerns explanations from another source that follow the person's own response and say what the evidence warrants.

The second principle comes from research on learning: what is retained depends on the processing done during practice. Working an answer out teaches more than receiving it (Bertsch et al., 2007; Chi \& Wylie, 2014), attempts that fail and are followed by the answer or by instruction also teach, though not always as much as attempts that succeed (Kapur, 2014; Kornell et al., 2009; Metcalfe, 2017), and students who click through a tutor's hints to reach the answer learn less (Aleven et al., 2003; Baker et al., 2004). Effort spent on parts of a task that are irrelevant to what is to be learned does not help (Sweller et al., 2019), and processing of meaning is retained better than shallow processing (Craik \& Lockhart, 1972). The intention to learn adds little once the processing is fixed: in classic studies of incidental learning, people who processed words in the same way remembered about as much whether or not they had been told to learn them (Hyde \& Jenkins, 1973). Worked examples teach novices well, most of all when learners explain the steps to themselves (Chi et al., 1989; Sweller et al., 2019). What is learned carries over to a test to the extent that the processing the test requires matches the processing done in practice (Morris et al., 1977). Performance during practice is an unreliable guide to all of this, which shows only in later tests, and practice that is easy and accurate can teach less than practice that is harder (Schmidt \& Bjork, 1992; Soderstrom \& Bjork, 2015).

The third principle comes from research on memory. Each use of knowledge strengthens it, the strength fades between uses, and uses spaced over time build strength that lasts (Cepeda et al., 2008; Roediger \& Karpicke, 2006). The activation model of memory describes both effects (Anderson et al., 2004; Pavlik \& Anderson, 2005). Retention is not the whole of transfer: people seldom notice unprompted that a new case calls for a rule they know (Gick \& Holyoak, 1983), although a statistical rule taught in one domain has been used in another (Fong \& Nisbett, 1991).

For claims in particular, corrections, accuracy prompts, media-literacy training and inoculation improve how people respond to false claims (Guess et al., 2020; Pennycook et al., 2021; Roozenbeek et al., 2022; Walter et al., 2020), but where durability has been measured the effects faded within weeks or months (Ecker et al., 2022; Maertens et al., 2021); rules for assessing claims about treatments, taught in school lessons or by a podcast, still helped a year later (Nsangi et al., 2020; Semakula et al., 2020). Together, these literatures do not yet establish how perceived reliability changes checking during practice and thereby supports delayed, rule-specific transfer.

\textbf{The new claim.} VLT concerns practice in which every claim is followed by a correct explanation, from a source other than the assistant, that says what the evidence warrants and names the rule that decides it. It combines established accounts of reliance on automation, learning through generation, and retention, and holds that people then learn to judge claims mainly from the checks they make themselves. A check-requiring task makes the check certain, and an assistant that seems fallible, because people have met its errors or have been told how often it errs, makes it more likely. Its distinctive claim is that, with correct rule-specific feedback held constant, task demands and perceived fallibility can substitute for one another because both prompt the same learning-relevant act: checking the evidence. The strict model assumes that once a genuine check occurs, its learning value does not depend on what prompted it. This restriction is most plausible for short, well-defined checks that learners can already attempt. It is less plausible when doubt changes the depth of reasoning or attention to subsequent feedback. The proposed tests therefore ask whether reliability has effects on retained learning beyond those attributable to checking.

\textbf{What is learned.} I distinguish four recurring ways in which claims go beyond their evidence, a classification drawn from analyses of health news (Schwitzer, 2008; Sumner et al., 2014). They infer causation from an association (Waldmann, 2017), extend a finding to people who were not studied, treat a proxy measure as the outcome that matters, or extend an effect beyond the dose, duration or setting tested. Each has a rule that detects it, such as `a finding may not hold for people unlike those studied'. A rule in this sense is a unit of knowledge that says when a check applies (Koedinger et al., 2012). Because it concerns how evidence bears on a claim, and not the claim's topic, a rule learned on one topic can be applied to another. A rule can also show that evidence does bear on a claim: when the people studied are those the claim is about, the evidence passes this check. The three warranted responses follow from the checks. A claim should be accepted when the evidence that passes every check that applies supports it, rejected when that evidence contradicts it, and withheld when no evidence passes the checks. Table 1 sets out one family of practice claims on the population rule, built so that the evidence meets the other three rules and this rule alone decides the response.

\begin{table}[tbp]
\caption{One family of practice claims on the population rule, with a test claim on a new topic}
\small\setstretch{1.0}
\begin{tabular}{@{}>{\raggedright\arraybackslash}p{0.175\linewidth}@{\hspace{1.4em}}>{\raggedright\arraybackslash}p{0.775\linewidth}@{}}
\toprule
Element & Example \\
\midrule
Rule & Population: a finding may not hold for people unlike those studied. \\ \addlinespace[3pt]
Claim & A sports drink improves the performance of adults who do not exercise regularly. \\ \addlinespace[3pt]
Evidence that warrants accepting & A randomised trial in adults who do not exercise regularly found that those given the drink performed better than those given a placebo drink. The people studied are those the claim is about, and the evidence supports the claim. \\ \addlinespace[3pt]
Evidence that warrants rejecting & A randomised trial in adults who do not exercise regularly found that those given the drink performed worse than those given a placebo drink; a second trial, in trained athletes, found that it improved their performance. Only the first trial bears on the claim, and it contradicts the claim. \\ \addlinespace[3pt]
Evidence that warrants withholding judgment & A randomised trial in trained athletes found that those given the drink performed better. No evidence concerns the people the claim is about. \\ \addlinespace[3pt]
Flip question & Which of these changes in how the evidence was obtained, each leaving every finding as it is, would change the response the evidence warrants? Options for the withholding version: the trial was run in adults who do not exercise regularly (correct: accept); the trial was run in younger trained athletes (same feature, response unchanged); the trial lasted ten weeks instead of eight; performance was measured on a cycling test instead of a running test. The last two change features that the rules on duration and proxy outcomes concern without breaking them. The accepting and rejecting versions offer the same kinds of options. \\ \addlinespace[3pt]
Explanation, withholding version & The evidence is insufficient to judge the claim. The trial studied trained athletes, and the claim is about adults who do not exercise regularly. Population rule: a finding may not hold for people unlike those studied. \\ \addlinespace[3pt]
Test claim on a new topic & A reading app improves reading in adults learning a second language; the only randomised trial, in children learning a second language, found that it improved their second-language reading. Warranted response: withhold judgment. \\ \addlinespace[3pt]
Held fixed by construction & Every trial is randomised, measures the performance the claim names and gives the drink at the dose and in the setting the claim describes, so the rules on causation, proxy outcomes and dose and setting are met and the population rule alone decides the response. \\
\bottomrule
\end{tabular}
\end{table}

\textbf{Relation to a broader framework.} A related cognitive-work-allocation framework distinguishes the operations people perform with AI and the capability trajectories that may follow (Trattner, 2026b). VLT develops a specific process account within that broader framework: it links perceived reliability to evidence checking, and the timing and distribution of checks to retained rule use. In that framework's terms, copying a verdict is only nominal participation in the operation of judging, and feedback on the criterion that decides a claim is what can make a corrected attempt useful practice. The contribution of VLT is the formal restriction and resulting experimental predictions, rather than the general distinction between assisted performance and independent capability. It also builds on a protocol for evaluating what AI-assisted verification leaves for unaided judgment (Trattner, 2026a).

\hypertarget{the-process-model}{%
\section{3 The process model}\label{the-process-model}}

Figure 1 shows the process that VLT proposes. The formal model states it in four components: belief updating, checking, learning and retention, and response selection. For each assumption I give the reason for it and the observation that would show it to be wrong.

\begin{figure}
\hypertarget{fig1}{%
\centering
\includegraphics{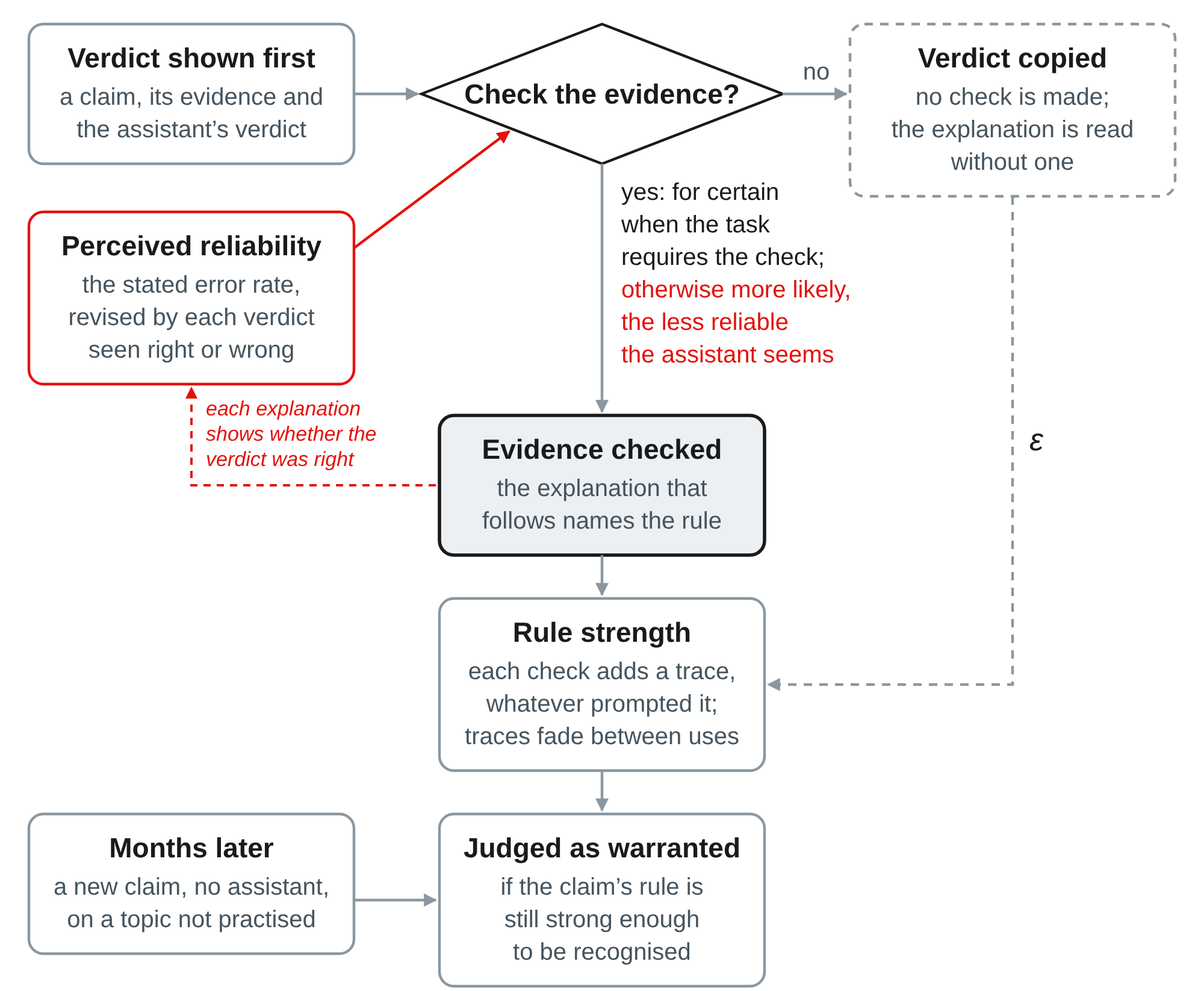}
\caption{The process Verification Learning Theory proposes. Perceived reliability starts from the stated error rate and is revised by every explanation, which shows whether the verdict was right. A person checks the evidence for certain when the task requires the check, and otherwise is more likely to check it the less reliable the assistant seems. Each check, successful or not, adds strength to the rule the claim calls for, whatever prompted it, while an explanation read without a check adds only a small fraction \(\varepsilon\) of that strength. Months later, a new claim is judged as its evidence warrants if the rule it calls for is still strong enough to be recognised.}\label{fig1}
}
\end{figure}

\textbf{Belief updating.} A person told that the assistant errs on a given share of claims starts from the corresponding reliability \(\rho_{0}\), one minus that share; a person told that it is always right starts from \(\rho_{0}\) = 1. Every explanation shows whether the verdict was right, and when person \(i\) meets claim \(k\), having seen \(n\) verdicts of which \(m\) proved right, their perceived reliability is

\begin{equation}
\hat{\rho}_{ik} = \frac{\kappa_{i}\,\rho_{0} + m_{ik}}{\kappa_{i} + n_{ik}}
\end{equation}

\noindent a weighted average of the stated reliability and the share of verdicts seen to be right, in which the statement counts as much as \(\kappa\) verdicts. It is the simplest way of combining what people are told about an aid's accuracy with the accuracy they observe, which people do (Yin et al., 2019). Errors met lower \(\hat{\rho}\) and correct verdicts raise it, so a stated error rate that is never borne out loses force with practice, and faster when \(\kappa\) is small. The component fails if ratings of reliability do not move towards the share of verdicts seen to be right. If a stated error rate does not lower perceived reliability in the first place, the component has not failed, but a test of its consequences has not met its conditions.

\textbf{Checking.} Let \(T\) = 1 if the task on a practice claim is check-requiring and \(T\) = 0 if it is copyable. The probability that person \(i\) checks the evidence for claim \(k\) is

\begin{equation}
q_{ik} = T_{ik} + (1 - T_{ik})\, h(\hat{\rho}_{ik};\, c_{i},\, \ell_{ik})
\end{equation}

\noindent where \(h\), between 0 and 1, falls as perceived reliability rises and as the effort \(\ell\) that the check takes rises, and rises with the person's self-confidence \(c\), their confidence in their own ability to evaluate evidence. The reason is that people examine an aid's output when they doubt it or trust their own judgment more than the aid (Lee \& Moray, 1994; Lee \& See, 2004), and avoid mental effort when they can (Kool et al., 2010). With a copyable task, the check is the person's choice. A check-requiring task makes it certain for anyone who answers in earnest, because a correct response cannot be produced without it, so the assistant's reliability cannot change whether a check is made. This is the first assumption of the strict model, \emph{compliance}: people answer a check-requiring task in earnest whatever the assistant. It fails if the share of answers that are guesses, or attempts at another rule, differs between reliability conditions. The component as a whole fails if people who rate the assistant as less reliable do not check more with a copyable task. An effort-matched task that neither requires the check nor makes it cheaper or costlier, for instance by drawing attention to the rule's condition, has the same \(q\) as the verdict task.

\textbf{Learning and retention.} Every practice claim ends with the explanation, which names the rule the claim calls for and shows whether the verdict was right. If the person made the check, successfully or not, the claim adds to that rule a trace, a unit of strength laid down by one use of the rule, of weight \(w\) = 1. If not, reading the explanation adds a trace of weight \(w\) = \(\varepsilon\), which VLT holds to be well below 1. That is least plausible for novices, who learn well from studying worked examples (Kalyuga et al., 2003; Sweller et al., 2019); where \(\varepsilon\) is large, every gain the model predicts shrinks (Section 5). Counting a failed check in full is a simplification, since attempts that fail and are then corrected teach, but not always as much as attempts that succeed (Metcalfe, 2017). In the model, giving failed checks half the weight leaves the effect of a stated error rate at zero when the task requires the check. It makes an erring assistant teach slightly less than an always-right one there, because wrong verdicts make checks fail more often (Supplementary Material S1). Two further assumptions complete the strict model. The second, \emph{equal learning per check}, is that every check adds a trace of the same weight, whatever prompted it, and that the explanation that follows is read as closely after one check as after another. Its reason is the one given by research on incidental learning: what is learned is set by the processing done, here the check of the rule's condition and the explanation that follows, and not by the motive for doing it (Hyde \& Jenkins, 1973). The third assumption is that an explanation read without a check teaches equally in every condition, so that \(\varepsilon\) does not depend on the task or the assistant. Its reason is that the explanation comes from an independent source and is the same in every condition; it fails if the time spent on explanations read without a check differs between conditions.

Equal learning per check is the strict model's riskiest assumption. Doubt might make a check more thorough, sharpen attention to the explanation, or make a correction more surprising and so better remembered. A \emph{graded form} of the model lets the weight of a check rise with its depth, measured independently of the outcome by process indicators such as the time spent on the part of the evidence the rule concerns. It is needed if such indicators forecast later judgment beyond the checks themselves. If the assistant's reliability also changes those indicators, or delayed unaided judgment, when the task requires the check, the depth account of Section 4 is needed instead.

Each trace fades between uses. The strength of rule \(s\) for person \(i\) at time \(t\) is

\begin{equation}
A_{is}(t) = \ln\Big( e^{\beta_{is}} + \sum_{j} w_{isj}\,(t - t_{isj})^{-d_{isj}} \Big), \qquad d_{isj} = \delta\, e^{A_{is}(t_{isj}^{-})} + \alpha
\end{equation}

\noindent where \(\beta\) is the rule's strength before practice, which a pretest can estimate, the sum runs over the traces laid down so far, and the \(j\)-th trace was laid down at time \(t_{isj}\) with weight \(w_{isj}\) and decays at rate \(d_{isj}\). As in Pavlik and Anderson (2005), the decay rate of a trace rises with the rule's strength just before the trace is laid down, with \(\alpha\) and \(\delta\) as decay parameters, which is what makes spaced checks last longer than massed ones. Time \(t\) is psychological time, in which time after a session counts \(\psi\) times as much as time within it, with \(\psi\) below 1, as models of this kind need in order to fit retention over long intervals (Anderson et al., 1999). With a fixed decay rate, the equation is a variant of the base-level learning equation of ACT-R (Anderson et al., 2004), which has predicted which tags people reuse in social bookmarking systems (Trattner et al., 2016) but, as far as I know, has not been applied to rules for judging evidence. The retention component fails if, with the checks measured validly and the spacing of practice varying between people, the model forecasts later errors no better than simpler models of the same checks, such as their number and the time since the last one.

\textbf{Response selection.} The model assigns each new claim the rule it calls for. It therefore predicts retention and use of a specified rule; it does not explain how people identify which rule applies when an unfamiliar claim offers ambiguous or competing cues, where the surface of a new topic can hide the rule (Gick \& Holyoak, 1983). The strength of the rule that a new claim \(k\) calls for sets how readily the person recognises that the rule applies, which the activation model expresses as a retrieval probability (Anderson et al., 2004):

\begin{equation}
\pi_{is}(t) = \frac{1}{1 + e^{-(A_{is}(t) - \tau)/\sigma}}
\end{equation}

\noindent where \(\tau\) is a threshold and \(\sigma\) the noise in strength. The person accepts, rejects or withholds judgment with probabilities

\begin{equation}
P(r \mid i, k) = \frac{e^{U_{ikr}}}{\sum_{r'} e^{U_{ikr'}}}, \qquad U_{ikr} = b_{ir} + \gamma_{ikr} + \lambda\, \pi_{i s(k)}(t)\, \mathbf{1}_{k}(r)
\end{equation}

\noindent for each response \(r\), where \(s\)(\(k\)) is the rule the claim calls for and the indicator in the last term equals 1 if \(r\) is the response that the evidence of claim \(k\) warrants and 0 otherwise. Recognition adds a pull of size \(\lambda\)\(\pi\) to the warranted response alone; \(b\) is the person's general tendency towards each response, and \(\gamma\) the pull of the claim itself, for example through how plausible or familiar it sounds. The component separates rule use from response tendency only under two identification assumptions: the tendency \(b\) is the same for every claim, whatever rule it calls for and whatever response it warrants, and a learned rule changes the responses only to claims that call for it. A change that appears equally on claims calling for an unpractised rule, or that favours one response on every claim, falls outside what the component attributes to learning.

\textbf{Outcome and measurement.} Delayed unaided judgment is scored as balanced accuracy: the mean, over the three warranted responses, of the share of new claims judged correctly, so that a person who gives one response to every claim scores one third. Checks themselves are not observed. With a check-requiring task, the response shows an attempt unless it was a guess; with a copyable task the evidence is indirect, such as the time spent on the evidence after the verdict or an override of a wrong verdict. Because the indicators differ between tasks and are noisy, any use of inferred checks needs a measurement model. Supplementary Material S2 proposes one; it has not been validated.

\hypertarget{predictions-and-closest-alternatives}{%
\section{4 Predictions and closest alternatives}\label{predictions-and-closest-alternatives}}

\textbf{Predictions.} The predictions concern comparisons in which the claims, the evidence and a correct explanation after every claim are held constant, and only the task and the assistant's reliability, or what people are told about it, vary. Six practice conditions cross the task, the verdict task (R) or the flip task (X), with three assistants: one that is always right and is said to be (E); one said to err at a stated rate that never errs (B, for belief); and one that errs at that rate and is said to (F). A seventh condition (M) pairs an always-right assistant with a copyable task that takes as long as the flip task without requiring the check. The outcome is delayed unaided judgment months later, on new claims that call for the practised rules or for an unpractised one.

\begin{enumerate}
\def\labelenumi{\arabic{enumi}.}
\tightlist
\item
  \textbf{Task gain.} E-X produces better delayed unaided judgment than E-R, because the flip task makes the check certain.
\item
  \textbf{Belief gain.} B-R produces better delayed unaided judgment than E-R, because a check is prompted by what people believe about the assistant, whether or not they see it err. Since every correct verdict raises perceived reliability, the extra checks should decline over practice, and the gain should be smaller than the fallibility gain of prediction 4. This is VLT's riskiest directional prediction, since with automated aids a warning has protected people less than the experience of a failure (Bahner et al., 2008).
\item
  \textbf{No belief effect once checking is required.} B-X and E-X differ by no more than a margin fixed in advance, because the check is already certain. With prediction 2, this implies a task-by-belief interaction. The verdicts and explanations are identical in E-R, B-R, E-X and B-X, so predictions 2 and 3 are the cleanest test of the account.
\item
  \textbf{Fallibility gain.} F-R produces better delayed unaided judgment than E-R, and better than B-R, because the errors met keep doubt alive. F-X and E-X differ by no more than the margin.
\item
  \textbf{Effort alone does not teach.} M and E-R differ by no more than the margin, provided that answering M neither requires the check nor makes it cheaper or costlier (Sweller et al., 2019).
\item
  \textbf{Selectivity.} Each gain appears on claims that call for practised rules and not on claims that call for an unpractised one, and on all three warranted responses, with no loss of accuracy on claims that warrant acceptance or on claims whose evidence passes the check only narrowly.
\item
  \textbf{Checks forecast retention.} People who started with the same prior knowledge and made the same checks at the same times hold the same rules, whatever prompted the checks. A memory model of each person's checks, fixed before the outcome is seen, should therefore forecast which kinds of claims that person will misjudge, and the condition in which they practised should add no more than a margin to that forecast. This prediction is a prospective extension, testable only once checks can be measured with a validated measurement model (Section 6).
\end{enumerate}

Predictions 3 and 5, the second part of prediction 4, the absence of a gain on the unpractised rule and the absence of a condition effect in prediction 7 follow from the strict model's definitions, and the signs of the gains in predictions 1, 2 and 4 follow from its structure. Section 5 separates these from the sizes of the gains, which depend on the parameters. Table 2 marks which predictions each alternative account shares.

\textbf{The closest alternative: depth and feedback encoding.} The account closest to VLT holds that doubt about the assistant also changes how much is learned from a check or from the explanation that follows. Processing of meaning is retained better than shallow processing (Craik \& Lockhart, 1972), feedback that contradicts an answer given with high confidence is corrected especially well (Butterfield \& Metcalfe, 2001; Metcalfe, 2017), and surprising feedback is better remembered (Fazio \& Marsh, 2009). People who doubt the assistant may therefore examine the evidence more thoroughly, attend more closely to the explanation, or remember a correction better because it surprised them. In the model, this means letting the weight of a trace rise with doubt. Without such an extra effect the account reduces to strict VLT, which it therefore nests.

The two accounts predict the same pattern wherever checking is the person's choice, although the depth account predicts larger gains. With the copyable task, both predict a belief gain and a larger fallibility gain, both expect effort alone to teach nothing, and both expect the gains to be specific to practised rules (Figure 2A and 2B). In delayed unaided judgment, they diverge only where the task already requires the check. Requiring a check removes the effect of doubt on learning if doubt acts only by deciding whether a check is made: once the task fixes the check, and the check is a short, well-defined examination that learners can already perform, there is nothing left for doubt to change. It does not remove that effect if doubt also changes how thoroughly the evidence is processed or how closely the explanation is read; then B-X and F-X should exceed E-X. Both accounts predict the task-by-belief interaction, so only the effect of belief when checking is required separates them.

That separation has limited resolution. In the simulation of Section 5, letting doubt raise the learning value of every check and explanation by half, for a person whose perceived reliability equals the stated rate, left B-X a median of 0.5 points of balanced accuracy above E-X, against a belief gain of 3.7 points with the copyable task. Doubling that learning value left a median of 0.9 points, and tripling it 1.4 points (Figure 3C). The effect is small for two reasons: once every claim is checked, the rules are already strong, and the doubt of people told of an error rate fades as they see correct verdicts. A broader depth or feedback-encoding model can therefore reproduce strict VLT when its extra effects are zero and come close to it when they are small. Finding a negligible effect of belief when checking is required can support the simpler restriction within an informative tolerance, but it cannot uniquely establish the mechanism, and a test of realistic size can rule out only effects larger than its margin (Section 6). Practice that keeps E-X further below its ceiling sharpens the test: with three practice claims per rule instead of eight, the same doubling left 2.2 points. If the flip task already demands the deepest processing people are capable of, the depth account also predicts no effect of belief, since doubt has no depth left to add; a null result then favours VLT only where measured depth is below its ceiling. Measures of depth separate all three accounts: the strict form predicts that they add nothing to a forecast once the checks are known; the graded form, that they add to it but do not differ between reliability conditions when the task requires the check; and the depth account, that they differ.

\textbf{Errors met when checking is required.} The comparison of F-X with E-X is less clean than that of B-X with E-X. A flip question presupposes the warranted response, so after a right verdict a person can take that response from the verdict and search only for the decisive feature, whereas after a wrong verdict they must first find the error. Error trials therefore ask for more work, and F-X can exceed E-X for that reason alone, as it can if errors teach by a route of their own (the two-route account in Table 2). Asking everyone for their own judgment before the flip response reduces this asymmetry but cannot remove it, since a right verdict can still be copied into the judgment. In the other direction, if failed checks teach less than successful ones, wrong verdicts that make checks fail more often push F-X slightly below E-X. B-X has no error trials, which is why predictions 2 and 3, not prediction 4, carry the test.

\textbf{Response tendencies.} A third account holds that a fallible assistant changes performance through response tendencies without improving the use of any rule: people who have met or been told of errors become warier and reject or withhold judgment more often. Caution of this kind raises accuracy on claims that warrant rejecting or withholding at the cost of claims that warrant acceptance, as when an intervention made people sceptical of accurate and inaccurate news alike (Lyons et al., 2024). Balanced accuracy removes the advantage of giving one response to every claim, but it does not by itself separate learning from caution: if people lean towards accepting before practice, caution alone can raise it. The model's response component separates the two only under its identification assumptions (Section 3), so a test should report the full matrix of responses by warranted response, separately for practised and unpractised rules. In the simulation, a learned rule pulls only towards the warranted response and response tendencies are symmetric, so by construction the fallibility gain appeared equally on all three warranted responses for practised rules and not at all for the unpractised rule. A shift of caution of half a unit on the log-odds scale, with learning unchanged, instead lowered accuracy on claims that warrant acceptance by 9.8 points and raised it on the other two by 4.3 points each, almost equally for practised and unpractised rules (Figure 2C). Signal-detection analyses separate discernment from bias in the same spirit for two responses (Batailler et al., 2022).

\textbf{Benchmark accounts.} Six further accounts serve as simple, transparent benchmarks (Table 2). Each is a simple model of one idea, not a statement of the research tradition it draws on. In \emph{answer retrieval}, people store the verdicts they saw, so nothing carries over to new claims. In \emph{explanation-driven learning}, every explanation teaches as much as a check, which is VLT with \(\varepsilon\) = 1. In \emph{task-only engagement}, only a task's demand to generate the answer teaches (Chi \& Wylie, 2014), which is VLT with checking unaffected by reliability. In \emph{error-driven learning}, only the contrast between a wrong verdict and its correction teaches (Metcalfe, 2017). In the \emph{two-route account}, the task and the errors met teach separately, their effects adding on the log-odds of a correct judgment. In \emph{general engagement}, a demanding task and doubt both raise effort, which teaches with diminishing returns whatever it is spent on. Because the two-route account adds effects on the log-odds scale, it too predicts a smaller fallibility gain where accuracy is already high, so a smaller gain under the flip task cannot by itself favour VLT (Loftus, 1978): VLT predicts a difference of zero, and a belief gain that the two-route account denies.

\begin{table}[tbp]
\caption{Predicted differences between practice conditions under strict VLT and the alternative accounts}
\footnotesize\setstretch{1.0}
\begin{tabular}{@{}>{\raggedright\arraybackslash}p{0.177\linewidth}@{\hspace{0.7em}}>{\raggedright\arraybackslash}p{0.072\linewidth}@{\hspace{0.7em}}>{\raggedright\arraybackslash}p{0.072\linewidth}@{\hspace{0.7em}}>{\raggedright\arraybackslash}p{0.092\linewidth}@{\hspace{0.7em}}>{\raggedright\arraybackslash}p{0.082\linewidth}@{\hspace{0.7em}}>{\raggedright\arraybackslash}p{0.092\linewidth}@{\hspace{0.7em}}>{\raggedright\arraybackslash}p{0.077\linewidth}@{\hspace{0.7em}}>{\raggedright\arraybackslash}p{0.102\linewidth}@{\hspace{0.7em}}>{\raggedright\arraybackslash}p{0.117\linewidth}@{}}
\toprule
Account & Task: E-X − E-R & Belief: B-R − E-R & Belief, check required: B-X − E-X & Fallibility: F-R − E-R & Fallibility, check required: F-X − E-X & Effort alone: M − E-R & Unpractised rule & Claims that warrant acceptance \\
\midrule
Strict VLT & + & + & 0 & + & 0 & 0 & 0 & no loss \\ \addlinespace[3pt]
Depth or feedback encoding (nests VLT) & + & + & + small & + & + small & 0 & 0 & no loss \\ \addlinespace[3pt]
Response tendencies & none & shift & shift & shift & shift & none & same shift & loss \\ \addlinespace[3pt]
Answer retrieval & 0 & 0 & 0 & 0 & 0 & 0 & 0 & no loss \\ \addlinespace[3pt]
Explanation-driven learning (VLT with \(\varepsilon\) = 1) & 0 & 0 & 0 & 0 & 0 & 0 & 0 & no loss \\ \addlinespace[3pt]
Task-only engagement (nested in VLT) & + & 0 & 0 & 0 & 0 & 0 & 0 & no loss \\ \addlinespace[3pt]
Error-driven learning & 0 & 0 & 0 & + & + & 0 & 0 & no loss \\ \addlinespace[3pt]
Two-route account & + & 0 & 0 & + & + & 0 & 0 & no loss \\ \addlinespace[3pt]
General engagement & + & + & + small & + & + small & + & + & no loss \\
\bottomrule
\end{tabular}
\par\smallskip\noindent\begin{minipage}{\linewidth}\footnotesize\setstretch{1.0}\emph{Note.} E: an assistant that is always right and said to be; B: said to err at a stated rate, never errs; F: errs at that rate; R: verdict task (copyable); X: flip task (check-requiring); M: an effort-matched copyable task with an assistant that is always right. +: better delayed unaided judgment; 0: no difference beyond a margin fixed in advance; small: smaller than with the copyable task; shift: more rejecting and withholding without better use of any rule, seen in the response matrix; none: no prediction. The last two columns give the gain on claims that call for an unpractised rule, and the change in accuracy on claims that warrant acceptance.\end{minipage}
\end{table}

\begin{figure}
\hypertarget{fig2}{%
\centering
\includegraphics{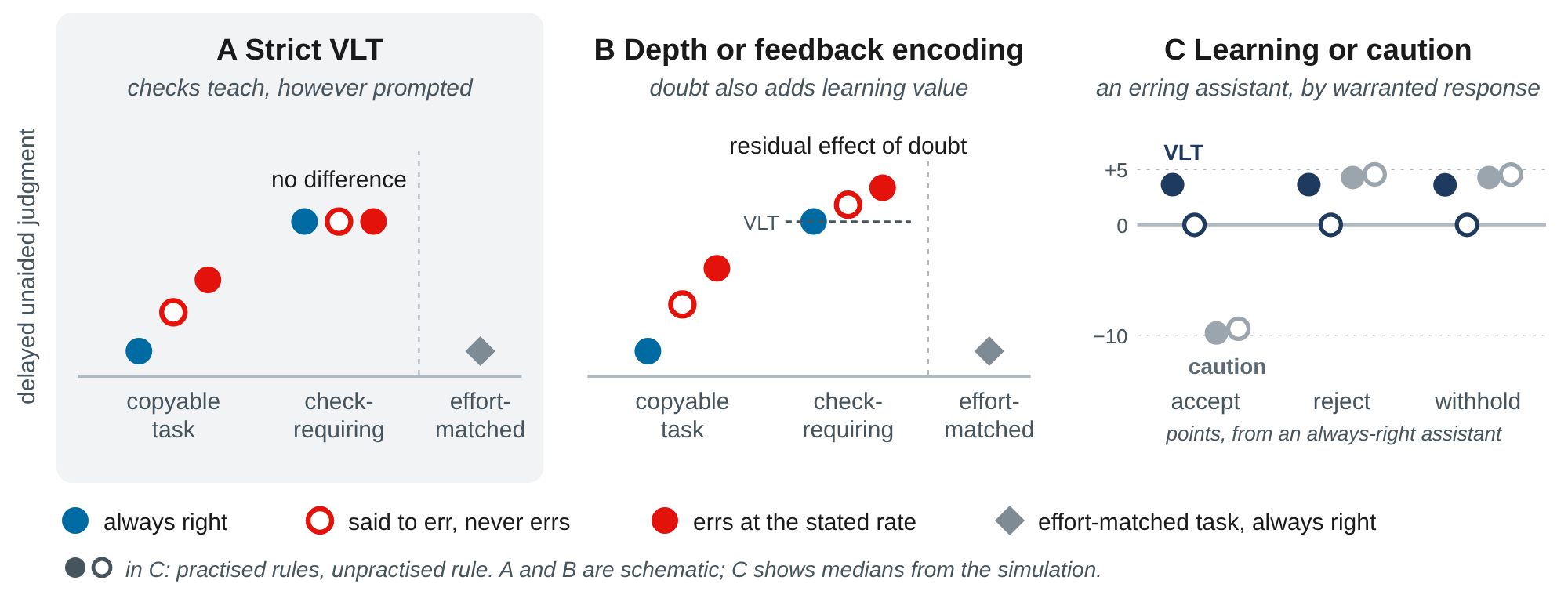}
\caption{Where strict VLT and its two closest alternatives agree and where they diverge. (A) Strict VLT: with the copyable task, a stated error rate and an assistant that errs improve delayed unaided judgment; with the check-requiring task the three assistants coincide; the effort-matched task equals the copyable task. (B) Depth or feedback encoding: with the copyable task the same order as VLT, with larger gains; with the check-requiring task doubt leaves a residual effect, and without it the account is strict VLT (dashed line). (C) Learning or caution: the change in accuracy from an always-right to an erring assistant, by the response that new claims warrant, on practised rules (filled) and an unpractised rule (open), under VLT and under an account in which the erring assistant leaves learning unchanged and lowers the tendency to accept by half a unit on the log-odds scale. Panels A and B are schematic; panel C shows medians over the 1,000 simulated parameter sets of Section 5.}\label{fig2}
}
\end{figure}

\hypertarget{computational-consequences}{%
\section{5 Computational consequences}\label{computational-consequences}}

To see which predictions follow from the model, and which depend on the values of its parameters, I simulated it with 1,000 sets of parameter values, drawn from ranges chosen to keep performance between chance and ceiling (Supplementary Material S1). Each set has 300 simulated people per condition, 24 practice claims on three rules in three weekly sessions, and a test 90 days later on new claims for the three practised rules and one unpractised rule. The simulation needs two assumptions that the model leaves open: \(h\) is a logistic function of perceived unreliability, self-confidence and effort, and during practice a person who does not check adopts the verdict. Because it computes each person's expected accuracy, and uses the same simulated people in every condition, its figures are the signs and sizes of the model's expected differences, not the rates at which a study would detect them. They are consequences of the model's assumptions and are no evidence that people learn as VLT says.

\textbf{What follows by construction.} Four patterns are identities of the strict model. With the check-requiring task the check is certain and every check counts the same, so B-X, E-X and F-X are equal, and the simulation returns differences of exactly zero. An effort-matched task that leaves the probability of a check unchanged teaches as much as the verdict task; nothing is gained on an unpractised rule; and once the checks are known, the condition adds nothing to the forecast. The signs of the three gains also follow from the model's structure. Because the same simulated people take part in every condition and \(h\) falls as perceived reliability rises, every claim checked with the always-right assistant is also checked with the other two, and every claim checked with the stated rate is also checked with the erring assistant. With \(\varepsilon\) below 1, the gains therefore cannot be negative, apart from a negligible effect of decay described below. These results show only that the code implements the definitions, and the simulation cannot test them.

\textbf{What depends on parameter choices.} The sizes of the gains depend on the parameters. Their medians were 5.8 points of balanced accuracy for the task gain, 3.6 for the fallibility gain and 2.4 for the belief gain (10th to 90th percentiles 1.3 to 13.9, 0.9 to 10.4 and 0.6 to 7.7). The share of claims checked was 100\% with the flip task, 57\% on average with the erring assistant, 40\% with the stated rate and 16\% with an assistant said to be always right. Four groups of parameters matter most (Figure 3). The first is learning from explanations alone: as \(\varepsilon\) rises, the gains shrink without reversing and vanish only as \(\varepsilon\) approaches 1, so the claim that \(\varepsilon\) is well below 1 can be tested only against a bound fixed in advance (Figure 3A). The second is belief updating. The belief gain was about half the fallibility gain when people gave the statement the weight of 2 to 10 verdicts, and about four-fifths when they gave it the weight of 20 to 30, because with little weight the share of claims checked fell faster over the three sessions (Figure 3B). The third is the learning value of a check under doubt. If people who have seen the assistant err take in less from every later explanation, an erring assistant teaches less than an always-right one even when the task requires the check, by a median of 1.4 points when the discount is a half. If doubt instead raises the learning value of a check, the effect under required checking is positive but small (Figure 3C). The fourth is retention. The gains grew between a test after 7 days and one after 90 days and changed little up to a year (Figure 3D), and with the check-requiring task, three sessions a week apart produced more than the same practice in one session (median 1.6 points), as the activation-dependent decay implies. The gains were also larger where rules were stronger before practice and more readily recognised (Supplementary Material S1). Replacing the logistic \(h\) by a linear or a threshold function changed the sizes of the gains but never reversed their signs.

Because a trace laid down while a rule is strong decays faster, an extra check need not raise a rule's later strength. Within the sampled ranges this is negligible: in a grid of more than 120,000 combinations of parameter values and schedules, an extra early check lowered later strength only when 15 or more checks followed it one second apart, and then by less than \(10^{-7}\) (Supplementary Material S1). The simulation also shows why accuracy during practice need not track learning. Within the copyable conditions, the share of claims a person checked was positively related to accuracy 90 days later (median correlations .22 to .42), whereas with an assistant said to be always right, accuracy during practice was negatively related to it (median −.31), because it came from copying.

\begin{figure}
\hypertarget{fig3}{%
\centering
\includegraphics{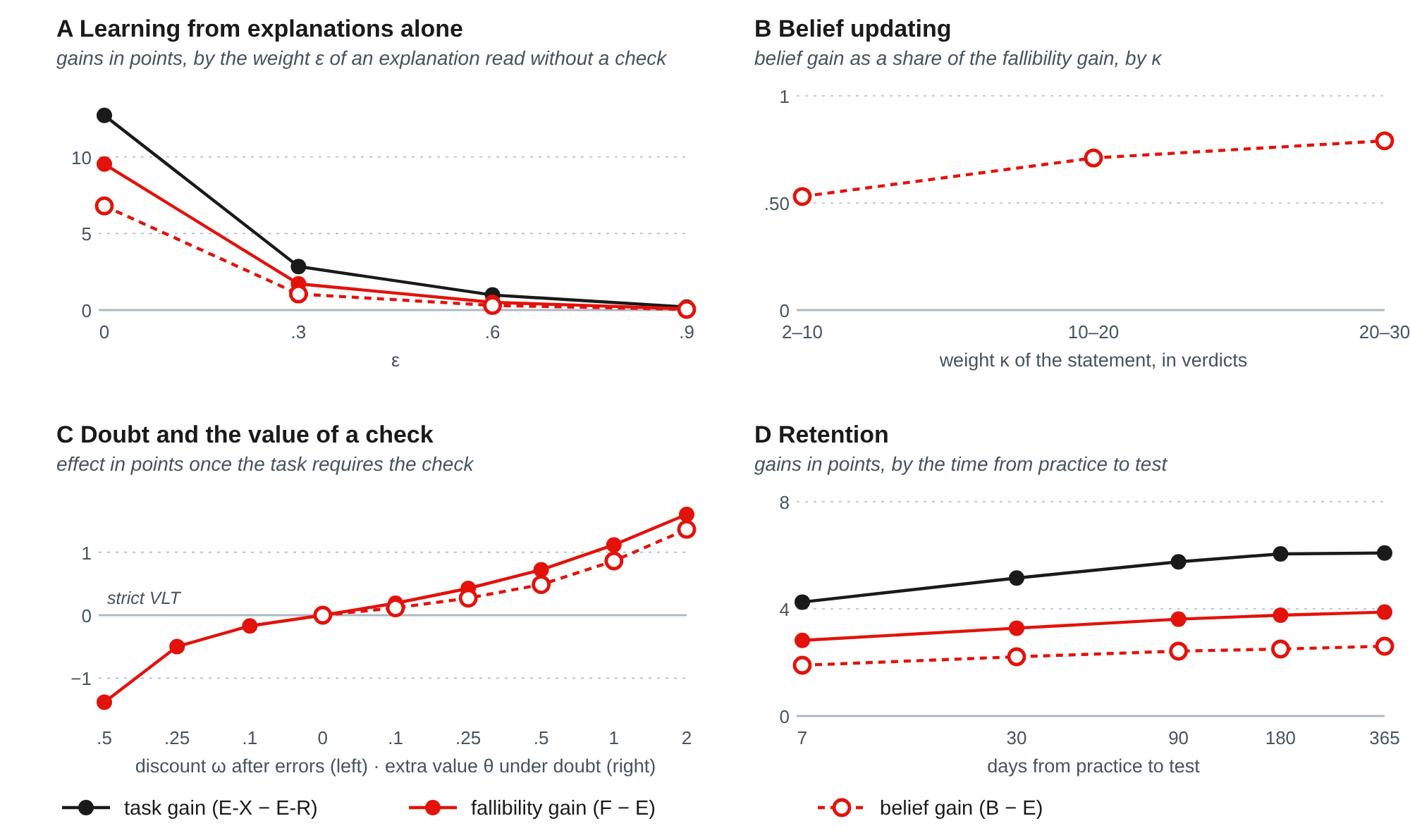}
\caption{How the simulated gains depend on the parameters that the model leaves open, in points of balanced accuracy 90 days after practice (medians over 1,000 parameter sets). (A) The task gain, the fallibility gain and the belief gain, with the weight \(\varepsilon\) of an explanation read without a check fixed at 0, .3, .6 and .9. (B) The belief gain as a share of the fallibility gain, by the weight \(\kappa\) that people give the statement. (C) The effect of the assistant when the task requires the check, if people who have seen it err take in less from every later explanation (discount ω, left), or if doubt raises the learning value of a check and of the explanation that follows (θ, right); zero is strict VLT. (D) The three gains when the test comes 7 to 365 days after practice.}\label{fig3}
}
\end{figure}

\textbf{What remains untested.} The simulation cannot show that people check, learn or forget as the model says. It did not compare the memory model with simpler models of the same checks, a comparison that discriminates only when the spacing of practice varies between people; it did not model how people identify the rule a claim calls for; and it did not simulate the measurement of checks.

\hypertarget{how-to-test-the-account}{%
\section{6 How to test the account}\label{how-to-test-the-account}}

\textbf{One design.} A test of the account needs the six conditions of Section 4 and the effort-matched condition M, with the same claims, evidence, explanations and practice schedule in all of them. The explanations come from a source other than the assistant, so that what people take from them does not depend on how far they trust it. The flip task offers a fixed set of changes, as in Table 1: one to the feature that the rule concerns that alters the warranted response, one to the same feature that does not, and changes to features that other rules concern that leave those rules satisfied. Only the first is correct, so a correct answer requires the check of the rule the claim calls for, and the second records an attempt at that rule that failed. Which rule is left unpractised should be counterbalanced across participants. Telling people in B that an assistant errs when it does not requires ethical approval and a debriefing.

The primary comparison is the two-by-two of E and B with each task. Its verdicts and explanations are identical, so any difference comes from what people were told and what the task asked. It tests the belief gain B-R − E-R as a directional effect against a smallest effect of interest, the effect of belief when checking is required, B-X − E-X, for equivalence, and the task-by-belief interaction. Restating the error rate at each session would keep the doubt of people in B alive, and fewer practice claims per rule would keep E-X further below its ceiling; both sharpen the test of prediction 3. The conditions with an erring assistant add the fallibility gain, its comparison with the belief gain, and F-X − E-X, read with the caution set out in Section 4. To limit the asymmetry of error trials, everyone should give their own judgment before the flip response, and the time spent on the evidence should be compared between error and correct trials in F-X, and between F-X and B-X. That procedure reduces the extra work of error trials and gives every condition the response format of the unaided test (Morris et al., 1977); it does not show that the processes are equivalent. A further condition with the same claims, evidence and explanations but no verdict at all would show whether practice with an assistant improves on practice without one.

\textbf{What must hold.} A result counts against VLT only if the conditions it presupposes were met, checked by measures fixed in advance and separate from the outcome (Table 3). Perceived reliability should be rated immediately after the statement, before the first verdict, and after each session: people in B and F must rate the assistant as less reliable than people in E, and VLT expects B's rating to recover as correct verdicts accumulate. A task instruction does not guarantee a genuine check, so engagement must be verified independently of whether the answer is correct. Flip answers are coded as attempts at the rule, attempts at another rule or guesses. A measure recorded before the answer, such as opening or dwelling on the part of the evidence the rule concerns, with a threshold for rapid guessing (Wise \& Kong, 2005), shows whether the evidence was examined. The shares of guesses and of attempts at another rule must be the same in every reliability condition. The effort-matched task must take as long as the flip task without requiring the check or drawing attention to any rule's condition, so the parts of the evidence must be the same, and unlabelled, in every condition. The time spent on the explanations should be recorded in every condition, since a fallible assistant that shortened it would count against the strict model. Delayed unaided judgment must be measured without the assistant, months after practice, on claims from topics never practised and on claims that call for an unpractised rule. These must include claims whose evidence passes the check only narrowly, as a trial in office workers who rarely exercise does for a claim about adults who do not exercise regularly, and the full matrix of responses should be reported.

\textbf{Margins and sizes.} Because VLT predicts negligible differences for prediction 3, for the second part of prediction 4 and for prediction 5, these can be tested only as equivalence, by showing that any difference is smaller than a margin fixed in advance. The margin should be an absolute difference justified independently of the data: the smallest difference in delayed unaided judgment that would matter for how people judge new claims, or, for prediction 3, the smallest effect of doubt that would make the depth account worth preferring. Setting it as a fraction of the observed gain with the copyable task would need a procedure for ratios, since that gain is itself estimated with error. The directional predictions likewise need smallest effects of interest fixed in advance, so that a gain too small to matter counts against them, and the numbers of people and claims should follow from these effects. Prediction 3 will demand the most. On a test of 18 claims for the practised rules, balanced accuracy in the simulation varied between people with a standard deviation of about 11 to 12 points once responses were sampled. An equivalence test with a margin of 2 points (\(\alpha\) = .05, power .80) then needs about 530 people per condition, and one with a margin of 1 point about 2,100. Given the effects in Figure 3C, a test of that size can rule out large effects of doubt under required checking, but not small ones. Table 3 sets out, for each prediction, the result that would count against strict VLT and what it would favour.

\begin{table}[tbp]
\caption{The tests of the design, the results that count against strict VLT, and the conditions under which they count}
\small\setstretch{1.0}
\begin{tabular}{@{}>{\raggedright\arraybackslash}p{0.130\linewidth}@{\hspace{1.4em}}>{\raggedright\arraybackslash}p{0.140\linewidth}@{\hspace{1.4em}}>{\raggedright\arraybackslash}p{0.205\linewidth}@{\hspace{1.4em}}>{\raggedright\arraybackslash}p{0.205\linewidth}@{\hspace{1.4em}}>{\raggedright\arraybackslash}p{0.195\linewidth}@{}}
\toprule
Prediction & Contrast & Result against strict VLT & What the result favours & Condition that must hold \\
\midrule
1 Task gain & E-X − E-R & No gain above the smallest effect of interest & Answer retrieval, explanation-driven or error-driven learning & Flip answers attempt the rule \\ \addlinespace[3pt]
2 Belief gain & B-R − E-R & No gain above the smallest effect of interest; or a gain as large as F-R − E-R although B's rating has recovered & No gain: error-driven learning or the two-route account. A gain as large: an effect of the statement that does not run through perceived reliability & B rates the assistant as less reliable than E from the first rating on \\ \addlinespace[3pt]
3 No belief effect, check required & B-X − E-X & A difference beyond the margin & Depth or feedback encoding if positive; a cost of doubt to the processing of explanations (not modelled) if negative & B rates the assistant as less reliable than E; equal shares of guesses and of attempts at another rule in B-X and E-X; measured depth below its ceiling \\ \addlinespace[3pt]
4 Fallibility gain & F-R − E-R, F-R − B-R, F-X − E-X & No gain in F-R − E-R or F-R − B-R; or F-X beyond the margin & F-X higher: the two-route account, extra work on error trials or depth. F-X lower: discounting of explanations after errors, or failed checks teaching less & F rates the assistant as less reliable; own judgment before the flip response \\ \addlinespace[3pt]
5 Effort alone & M − E-R & A difference beyond the margin & General engagement if M is higher & M takes as long as the flip task and leaves checking unchanged \\ \addlinespace[3pt]
6 Selectivity & Practised against unpractised rules; full response matrix & Gains as large on the unpractised rule; a loss on claims that warrant acceptance; a loss confined to narrowly passing claims on practised rules & General engagement; response tendencies; over-general rule learning, which the response component excludes by assumption & Test claims include claims that warrant acceptance and claims that pass only narrowly \\ \addlinespace[3pt]
7 Forecast (prospective) & Held-out score of the memory model with and without the condition & The condition improves the forecast beyond the margin; no gain over prior knowledge or accuracy during practice & The graded form, if measured depth absorbs the effect of the condition; the depth account, if depth itself differs between conditions & A validated measurement model; spacing that varies between people \\
\bottomrule
\end{tabular}
\end{table}

\textbf{Forecasting from checks.} Prediction 7 is a prospective extension, conditional on measuring checks validly. It needs a measurement model that infers checks from indicators recorded in the same way in every condition compared, allows for guessing and for error in each indicator, and has been validated against independent evidence of checking, such as think-aloud protocols. Supplementary Material S2 proposes such a model; it has not been validated. The memory model's parameters would be fixed on one group of people, with their checks inferred from all their data, and its forecasts scored on another group, with their checks inferred from practice data alone, by a scoring rule fixed in advance, such as the log probability of each response. The forecast should beat two sets of alternatives: prior knowledge, accuracy during practice and a general tendency to reject claims, which would show that checks matter; and simpler models of the same checks, which would support the account of retention. Under a fixed practice schedule, strength at the test depends mostly on the number of checks, so the spacing of practice must vary between people for the second comparison to discriminate. The condition should enter the forecast as a change in the weight of the traces, which can capture a difference in learning, and not only as a tendency towards each response, which captures caution alone. Even a successful forecast would not show that the task and the assistant act only through the checks unless no unmeasured variable, such as attention or motivation, affects both checking and learning; randomising the condition does not ensure this, because checking happens after randomisation (Bullock et al., 2010; Montgomery et al., 2018). Prediction 3 is the randomised counterpart, since the flip task fixes checking by design.

\textbf{Locating a failure.} When a prediction fails, these measures locate the step. An unchanged rating means that the test did not meet its conditions. The checking, acquisition and retention components fail in turn if lower perceived reliability does not raise checks, if checks do not raise strength by the end of practice, or if strength does not last; an unaided test at the end of practice, counted as practice in the model, separates the last two (Roediger \& Karpicke, 2006).

\hypertarget{scope-and-implications}{%
\section{7 Scope and implications}\label{scope-and-implications}}

\textbf{Scope.} VLT is stated for controlled rule use: judging claims against evidence that is given, in practice where every claim is followed by a correct explanation that names the rule it calls for. Three things lie beyond it. The first is broader transfer: how people recognise which rule an unfamiliar claim calls for when its cues are ambiguous or compete, and whether rules learned on health claims carry over to political claims, where identity and motivated reasoning work against applying a rule to claims that favour one's own side (Kahan et al., 2017). Transfer of this kind tests the account's reach, which the model does not yet describe; practising several rules mixed together, which helps people learn which rule applies (Rohrer, 2012), is one way to extend it. The second is practice without correct feedback, in which people cannot learn whether a verdict was right. The third is everyday oversight: whether judgment erodes under daily use of an assistant whose errors go unnoticed, and what happens to skills people already have, which the model, holding prior strength constant, does not address. Within its scope, the proposition should hold wherever an answer can be copied and correct feedback follows, in writing, programming, mathematics or clinical decisions, where research has documented costs of relying on AI (Bastani et al., 2025; Budzyń et al., 2025; Shen \& Tamkin, 2026); whether it does is an empirical question.

\textbf{Boundary conditions.} A check requires the knowledge to make it, and novices may gain little from an assistant's errors unless the errors are explained (Dieterich et al., 2025; Kalyuga et al., 2003); the explanation after every claim is meant to supply that knowledge. People may keep copying an assistant despite its errors (Shaw \& Nave, 2026); if they rate it as less reliable and still do not check, the checking component is wrong. People who have seen an algorithm err may stop relying on it (Dietvorst et al., 2015) and might also discount the explanations that follow its errors. Explanations from another source should limit this, and where it happens it makes a fallible assistant teach less even when the task requires the check (Figure 3C).

\textbf{Implications.} If the accuracy paradox holds, the assumption that a more reliable assistant is also the better teacher fails, at least where answers can be copied and correct explanations follow, as in training and tutoring, and accuracy during use cannot stand in for what users learn (Wiles et al., 2026). If a stated error rate teaches on its own, telling users specifically and credibly how often an assistant errs becomes a way of teaching. The generic notice that an assistant can make mistakes names no rate; if it acts like a stated rate given little weight, the model implies that its effect is small and fades wherever users see the assistant being right. If instead the task and the errors met teach separately, a demanding task and a fallible assistant are separate levers that designers can combine. Human oversight of AI presumes people who can judge an AI's output (European Parliament \& Council of the European Union, 2024), and people who judge well only while the assistant is present are least able to tell when it is wrong (Bainbridge, 1983). VLT holds that this ability is built through the checks that the use of an assistant asks for. If people check as often when they meet an assistant again, the model implies that those with better delayed judgment will override more of its wrong verdicts on claims that call for practised rules, without overriding more of its right ones. For the assistance dilemma (Koedinger \& Aleven, 2007), it suggests asking whether the help can be adopted without a check, as well as how much help to give. The model makes one question answerable: with feedback held constant, does it matter for what people retain why they checked the evidence, or only that they did?

\hypertarget{declarations}{%
\section{Declarations}\label{declarations}}

\textbf{Use of generative AI.} I developed the theory with the help of Claude (Anthropic), in the versions available on claude.ai in 2026, which I used to draft, critique and revise text. Claude also drafted this manuscript, its figures and the simulation code under my direction, and checked the references against my reference list. I decided the content of the theory and of this article, revised and approved the text, checked every reference against its source, and take full responsibility for the manuscript. Every number reported from the simulations was checked against the output of the code.

\textbf{Competing interests.} The author declares no competing interests.

\textbf{Data and code.} No data were collected or analysed for this article. The Python code for the simulations in Section 5 and Supplementary Material S1, with the outputs it writes, is available on the Open Science Framework at \href{https://osf.io/zd5j4/}{\url{https://osf.io/zd5j4/}}.

\textbf{Open practices statement.} No data were collected for this article and no study was preregistered: the article presents a theory and simulations of its formal model. The simulation code and its outputs are available at the address given under Data and code.

\section*{References}\addcontentsline{toc}{section}{References}

\begingroup\small\raggedright\setlength{\parindent}{-1.6em}\setlength{\leftskip}{1.6em}\setlength{\parskip}{2.5pt}\makeatletter\@afterindenttrue\makeatother

Aleven, V., Stahl, E., Schworm, S., Fischer, F., \& Wallace, R. (2003). Help seeking and help design in interactive learning environments. \emph{Review of Educational Research, 73}(3), 277--320. \url{https://doi.org/10.3102/00346543073003277}

Anderson, J. R., Bothell, D., Byrne, M. D., Douglass, S., Lebiere, C., \& Qin, Y. (2004). An integrated theory of the mind. \emph{Psychological Review, 111}(4), 1036--1060. \url{https://doi.org/10.1037/0033-295X.111.4.1036}

Anderson, J. R., Fincham, J. M., \& Douglass, S. (1999). Practice and retention: A unifying analysis. \emph{Journal of Experimental Psychology: Learning, Memory, and Cognition, 25}(5), 1120--1136. \url{https://doi.org/10.1037/0278-7393.25.5.1120}

Bahner, J. E., Hüper, A.-D., \& Manzey, D. (2008). Misuse of automated decision aids: Complacency, automation bias and the impact of training experience. \emph{International Journal of Human-Computer Studies, 66}(9), 688--699. \url{https://doi.org/10.1016/j.ijhcs.2008.06.001}

Bainbridge, L. (1983). Ironies of automation. \emph{Automatica, 19}(6), 775--779. \url{https://doi.org/10.1016/0005-1098(83)90046-8}

Baker, R. S., Corbett, A. T., Koedinger, K. R., \& Wagner, A. Z. (2004). Off-task behavior in the Cognitive Tutor classroom: When students ``game the system''. In \emph{Proceedings of the SIGCHI Conference on Human Factors in Computing Systems} (pp.~383--390). ACM. \url{https://doi.org/10.1145/985692.985741}

Bastani, H., Bastani, O., Sungu, A., Ge, H., Kabakcı, Ö., \& Mariman, R. (2025). Generative AI without guardrails can harm learning: Evidence from high school mathematics. \emph{Proceedings of the National Academy of Sciences, 122}(26), Article e2422633122. \url{https://doi.org/10.1073/pnas.2422633122}

Batailler, C., Brannon, S. M., Teas, P. E., \& Gawronski, B. (2022). A signal detection approach to understanding the identification of fake news. \emph{Perspectives on Psychological Science, 17}(1), 78--98. \url{https://doi.org/10.1177/1745691620986135}

Bertsch, S., Pesta, B. J., Wiscott, R., \& McDaniel, M. A. (2007). The generation effect: A meta-analytic review. \emph{Memory \& Cognition, 35}(2), 201--210. \url{https://doi.org/10.3758/BF03193441}

Buçinca, Z., Malaya, M. B., \& Gajos, K. Z. (2021). To trust or to think: Cognitive forcing functions can reduce overreliance on AI in AI-assisted decision-making. \emph{Proceedings of the ACM on Human-Computer Interaction, 5}(CSCW1), Article 188. \url{https://doi.org/10.1145/3449287}

Buçinca, Z., Swaroop, S., Paluch, A. E., Doshi-Velez, F., \& Gajos, K. Z. (2025). Contrastive explanations that anticipate human misconceptions can improve human decision-making skills. In \emph{Proceedings of the 2025 CHI Conference on Human Factors in Computing Systems}. ACM. \url{https://doi.org/10.1145/3706598.3713229}

Budzyń, K., Romańczyk, M., Kitala, D., Kołodziej, P., Bugajski, M., Adami, H. O., Blom, J., Buszkiewicz, M., Halvorsen, N., Hassan, C., Romańczyk, T., Holme, Ø., Jarus, K., Fielding, S., Kunar, M., Pellisé, M., Pilonis, N., Kamiński, M. F., Kalager, M., . . . Mori, Y. (2025). Endoscopist deskilling risk after exposure to artificial intelligence in colonoscopy: A multicentre, observational study. \emph{The Lancet Gastroenterology \& Hepatology, 10}(10), 896--903. \url{https://doi.org/10.1016/S2468-1253(25)00133-5}

Bullock, J. G., Green, D. P., \& Ha, S. E. (2010). Yes, but what's the mechanism? (Don't expect an easy answer). \emph{Journal of Personality and Social Psychology, 98}(4), 550--558. \url{https://doi.org/10.1037/a0018933}

Butterfield, B., \& Metcalfe, J. (2001). Errors committed with high confidence are hypercorrected. \emph{Journal of Experimental Psychology: Learning, Memory, and Cognition, 27}(6), 1491--1494. \url{https://doi.org/10.1037/0278-7393.27.6.1491}

Cepeda, N. J., Vul, E., Rohrer, D., Wixted, J. T., \& Pashler, H. (2008). Spacing effects in learning: A temporal ridgeline of optimal retention. \emph{Psychological Science, 19}(11), 1095--1102. \url{https://doi.org/10.1111/j.1467-9280.2008.02209.x}

Chatterji, A., Cunningham, T., Deming, D. J., Hitzig, Z., Ong, C., Shan, C. Y., \& Wadman, K. (2025). \emph{How people use ChatGPT} (NBER Working Paper No.~34255). National Bureau of Economic Research. \url{https://doi.org/10.3386/w34255}

Chi, M. T. H., Bassok, M., Lewis, M. W., Reimann, P., \& Glaser, R. (1989). Self-explanations: How students study and use examples in learning to solve problems. \emph{Cognitive Science, 13}(2), 145--182. \url{https://doi.org/10.1207/s15516709cog1302_1}

Chi, M. T. H., \& Wylie, R. (2014). The ICAP framework: Linking cognitive engagement to active learning outcomes. \emph{Educational Psychologist, 49}(4), 219--243. \url{https://doi.org/10.1080/00461520.2014.965823}

Craik, F. I. M., \& Lockhart, R. S. (1972). Levels of processing: A framework for memory research. \emph{Journal of Verbal Learning and Verbal Behavior, 11}(6), 671--684. \url{https://doi.org/10.1016/S0022-5371(72)80001-X}

Dieterich, S., Rumann, S., \& Rodemer, M. (2025). Conditions for effective learning from erroneous examples: A systematic review. \emph{Educational Psychology Review, 37}(4), Article 94. \url{https://doi.org/10.1007/s10648-025-10071-x}

Dietvorst, B. J., Simmons, J. P., \& Massey, C. (2015). Algorithm aversion: People erroneously avoid algorithms after seeing them err. \emph{Journal of Experimental Psychology: General, 144}(1), 114--126. \url{https://doi.org/10.1037/xge0000033}

Ecker, U. K. H., Lewandowsky, S., Cook, J., Schmid, P., Fazio, L. K., Brashier, N., Kendeou, P., Vraga, E. K., \& Amazeen, M. A. (2022). The psychological drivers of misinformation belief and its resistance to correction. \emph{Nature Reviews Psychology, 1}(1), 13--29. \url{https://doi.org/10.1038/s44159-021-00006-y}

European Parliament \& Council of the European Union. (2024). Regulation (EU) 2024/1689 laying down harmonised rules on artificial intelligence (Artificial Intelligence Act). \emph{Official Journal of the European Union}, L series, 12 July 2024.

Fazio, L. K., \& Marsh, E. J. (2009). Surprising feedback improves later memory. \emph{Psychonomic Bulletin \& Review, 16}(1), 88--92. \url{https://doi.org/10.3758/PBR.16.1.88}

Fong, G. T., \& Nisbett, R. E. (1991). Immediate and delayed transfer of training effects in statistical reasoning. \emph{Journal of Experimental Psychology: General, 120}(1), 34--45. \url{https://doi.org/10.1037/0096-3445.120.1.34}

Gajos, K. Z., \& Mamykina, L. (2022). Do people engage cognitively with AI? Impact of AI assistance on incidental learning. In \emph{Proceedings of the 27th International Conference on Intelligent User Interfaces} (pp.~794--806). ACM. \url{https://doi.org/10.1145/3490099.3511138}

Gick, M. L., \& Holyoak, K. J. (1983). Schema induction and analogical transfer. \emph{Cognitive Psychology, 15}(1), 1--38. \url{https://doi.org/10.1016/0010-0285(83)90002-6}

Guess, A. M., Lerner, M., Lyons, B., Montgomery, J. M., Nyhan, B., Reifler, J., \& Sircar, N. (2020). A digital media literacy intervention increases discernment between mainstream and false news in the United States and India. \emph{Proceedings of the National Academy of Sciences, 117}(27), 15536--15545. \url{https://doi.org/10.1073/pnas.1920498117}

Hyde, T. S., \& Jenkins, J. J. (1973). Recall for words as a function of semantic, graphic, and syntactic orienting tasks. \emph{Journal of Verbal Learning and Verbal Behavior, 12}(5), 471--480. \url{https://doi.org/10.1016/S0022-5371(73)80027-1}

Kahan, D. M., Peters, E., Dawson, E. C., \& Slovic, P. (2017). Motivated numeracy and enlightened self-government. \emph{Behavioural Public Policy, 1}(1), 54--86. \url{https://doi.org/10.1017/bpp.2016.2}

Kalyuga, S., Ayres, P., Chandler, P., \& Sweller, J. (2003). The expertise reversal effect. \emph{Educational Psychologist, 38}(1), 23--31. \url{https://doi.org/10.1207/S15326985EP3801_4}

Kapur, M. (2014). Productive failure in learning math. \emph{Cognitive Science, 38}(5), 1008--1022. \url{https://doi.org/10.1111/cogs.12107}

Koedinger, K. R., \& Aleven, V. (2007). Exploring the assistance dilemma in experiments with Cognitive Tutors. \emph{Educational Psychology Review, 19}(3), 239--264. \url{https://doi.org/10.1007/s10648-007-9049-0}

Koedinger, K. R., Corbett, A. T., \& Perfetti, C. (2012). The Knowledge-Learning-Instruction framework: Bridging the science-practice chasm to enhance robust student learning. \emph{Cognitive Science, 36}(5), 757--798. \url{https://doi.org/10.1111/j.1551-6709.2012.01245.x}

Kool, W., McGuire, J. T., Rosen, Z. B., \& Botvinick, M. M. (2010). Decision making and the avoidance of cognitive demand. \emph{Journal of Experimental Psychology: General, 139}(4), 665--682. \url{https://doi.org/10.1037/a0020198}

Kornell, N., Hays, M. J., \& Bjork, R. A. (2009). Unsuccessful retrieval attempts enhance subsequent learning. \emph{Journal of Experimental Psychology: Learning, Memory, and Cognition, 35}(4), 989--998. \url{https://doi.org/10.1037/a0015729}

Lee, J. D., \& Moray, N. (1994). Trust, self-confidence, and operators' adaptation to automation. \emph{International Journal of Human-Computer Studies, 40}(1), 153--184. \url{https://doi.org/10.1006/ijhc.1994.1007}

Lee, J. D., \& See, K. A. (2004). Trust in automation: Designing for appropriate reliance. \emph{Human Factors, 46}(1), 50--80. \url{https://doi.org/10.1518/hfes.46.1.50_30392}

Loftus, G. R. (1978). On interpretation of interactions. \emph{Memory \& Cognition, 6}(3), 312--319. \url{https://doi.org/10.3758/BF03197461}

Lyons, B., King, A. J., \& Kaphingst, K. A. (2024). A health media literacy intervention increases skepticism of both inaccurate and accurate cancer news among U.S. adults. \emph{Annals of Behavioral Medicine, 58}(12), 820--831. \url{https://doi.org/10.1093/abm/kaae054}

Maertens, R., Roozenbeek, J., Basol, M., \& van der Linden, S. (2021). Long-term effectiveness of inoculation against misinformation: Three longitudinal experiments. \emph{Journal of Experimental Psychology: Applied, 27}(1), 1--16. \url{https://doi.org/10.1037/xap0000315}

Metcalfe, J. (2017). Learning from errors. \emph{Annual Review of Psychology, 68}, 465--489. \url{https://doi.org/10.1146/annurev-psych-010416-044022}

Montgomery, J. M., Nyhan, B., \& Torres, M. (2018). How conditioning on posttreatment variables can ruin your experiment and what to do about it. \emph{American Journal of Political Science, 62}(3), 760--775. \url{https://doi.org/10.1111/ajps.12357}

Morris, C. D., Bransford, J. D., \& Franks, J. J. (1977). Levels of processing versus transfer appropriate processing. \emph{Journal of Verbal Learning and Verbal Behavior, 16}(5), 519--533. \url{https://doi.org/10.1016/S0022-5371(77)80016-9}

Nsangi, A., Semakula, D., Oxman, A. D., Austvoll-Dahlgren, A., Oxman, M., Rosenbaum, S., Morelli, A., Glenton, C., Lewin, S., Kaseje, M., Chalmers, I., Fretheim, A., Ding, Y., \& Sewankambo, N. K. (2020). Effects of the Informed Health Choices primary school intervention on the ability of children in Uganda to assess the reliability of claims about treatment effects, 1-year follow-up: A cluster-randomised trial. \emph{Trials, 21}(1), Article 27. \url{https://doi.org/10.1186/s13063-019-3960-9}

Parasuraman, R., \& Manzey, D. H. (2010). Complacency and bias in human use of automation: An attentional integration. \emph{Human Factors, 52}(3), 381--410. \url{https://doi.org/10.1177/0018720810376055}

Parasuraman, R., \& Riley, V. (1997). Humans and automation: Use, misuse, disuse, abuse. \emph{Human Factors, 39}(2), 230--253. \url{https://doi.org/10.1518/001872097778543886}

Pavlik, P. I., Jr., \& Anderson, J. R. (2005). Practice and forgetting effects on vocabulary memory: An activation-based model of the spacing effect. \emph{Cognitive Science, 29}(4), 559--586. \url{https://doi.org/10.1207/s15516709cog0000_14}

Pennycook, G., Epstein, Z., Mosleh, M., Arechar, A. A., Eckles, D., \& Rand, D. G. (2021). Shifting attention to accuracy can reduce misinformation online. \emph{Nature, 592}(7855), 590--595. \url{https://doi.org/10.1038/s41586-021-03344-2}

Rani, A., Danry, V., Liang, P. P., Lippman, A. B., \& Maes, P. (2026). Dialogues with AI reduce beliefs in misinformation but build no lasting discernment skills. In \emph{Proceedings of the 2026 CHI Conference on Human Factors in Computing Systems} (Article 792). ACM. \url{https://doi.org/10.1145/3772318.3790656}

Roediger, H. L., III, \& Karpicke, J. D. (2006). Test-enhanced learning: Taking memory tests improves long-term retention. \emph{Psychological Science, 17}(3), 249--255. \url{https://doi.org/10.1111/j.1467-9280.2006.01693.x}

Rohrer, D. (2012). Interleaving helps students distinguish among similar concepts. \emph{Educational Psychology Review, 24}(3), 355--367. \url{https://doi.org/10.1007/s10648-012-9201-3}

Roozenbeek, J., van der Linden, S., Goldberg, B., Rathje, S., \& Lewandowsky, S. (2022). Psychological inoculation improves resilience against misinformation on social media. \emph{Science Advances, 8}(34), Article eabo6254. \url{https://doi.org/10.1126/sciadv.abo6254}

Salomon, G., Perkins, D. N., \& Globerson, T. (1991). Partners in cognition: Extending human intelligence with intelligent technologies. \emph{Educational Researcher, 20}(3), 2--9. \url{https://doi.org/10.3102/0013189X020003002}

Schmidt, R. A., \& Bjork, R. A. (1992). New conceptualizations of practice: Common principles in three paradigms suggest new concepts for training. \emph{Psychological Science, 3}(4), 207--217. \url{https://doi.org/10.1111/j.1467-9280.1992.tb00029.x}

Schwitzer, G. (2008). How do US journalists cover treatments, tests, products, and procedures? An evaluation of 500 stories. \emph{PLoS Medicine, 5}(5), Article e95. \url{https://doi.org/10.1371/journal.pmed.0050095}

Semakula, D., Nsangi, A., Oxman, A. D., Oxman, M., Austvoll-Dahlgren, A., Rosenbaum, S., Morelli, A., Glenton, C., Lewin, S., Nyirazinyoye, L., Kaseje, M., Chalmers, I., Fretheim, A., Rose, C. J., \& Sewankambo, N. K. (2020). Effects of the Informed Health Choices podcast on the ability of parents of primary school children in Uganda to assess the trustworthiness of claims about treatment effects: One-year follow up of a randomised trial. \emph{Trials, 21}(1), Article 187. \url{https://doi.org/10.1186/s13063-020-4093-x}

Shaw, S. D., \& Nave, G. (2026). \emph{Thinking---fast, slow, and artificial: How AI is reshaping human reasoning and the rise of cognitive surrender} {[}Preprint{]}. SSRN. \url{https://doi.org/10.2139/ssrn.6097646}

Shen, J. H., \& Tamkin, A. (2026). \emph{How AI impacts skill formation} {[}Preprint{]}. arXiv. \url{https://doi.org/10.48550/arXiv.2601.20245}

Soderstrom, N. C., \& Bjork, R. A. (2015). Learning versus performance: An integrative review. \emph{Perspectives on Psychological Science, 10}(2), 176--199. \url{https://doi.org/10.1177/1745691615569000}

Sumner, P., Vivian-Griffiths, S., Boivin, J., Williams, A., Venetis, C. A., Davies, A., Ogden, J., Whelan, L., Hughes, B., Dalton, B., Boy, F., \& Chambers, C. D. (2014). The association between exaggeration in health related science news and academic press releases: Retrospective observational study. \emph{BMJ, 349}, Article g7015. \url{https://doi.org/10.1136/bmj.g7015}

Sweller, J., van Merriënboer, J. J. G., \& Paas, F. (2019). Cognitive architecture and instructional design: 20 years later. \emph{Educational Psychology Review, 31}(2), 261--292. \url{https://doi.org/10.1007/s10648-019-09465-5}

Toulmin, S. E. (1958). \emph{The uses of argument}. Cambridge University Press.

Trattner, C. (2026a). \emph{Epistemic transfer in AI-assisted verification: A framework and evaluation protocol} {[}Preprint{]}. arXiv. \url{https://doi.org/10.48550/arXiv.2608.08882}

Trattner, C. (2026b). \emph{Who does the thinking? How cognitive work allocation with AI shapes human capability} {[}Manuscript submitted for publication{]}.

Trattner, C., Kowald, D., Seitlinger, P., Kopeinik, S., \& Ley, T. (2016). Modeling activation processes in human memory to predict the use of tags in social bookmarking systems. \emph{The Journal of Web Science, 2}(1), 1--18. \url{https://doi.org/10.1561/106.00000004}

Waldmann, M. R. (Ed.). (2017). \emph{The Oxford handbook of causal reasoning}. Oxford University Press. \url{https://doi.org/10.1093/oxfordhb/9780199399550.001.0001}

Walter, N., Cohen, J., Holbert, R. L., \& Morag, Y. (2020). Fact-checking: A meta-analysis of what works and for whom. \emph{Political Communication, 37}(3), 350--375. \url{https://doi.org/10.1080/10584609.2019.1668894}

Wiles, E., Krayer, L., Abbadi, M., Awasthi, U., Kennedy, R., Mishkin, P., Sack, D., \& Candelon, F. (2026). Generative AI and the temporary upskilling of knowledge workers. \emph{Nature Human Behaviour}. Advance online publication. \url{https://doi.org/10.1038/s41562-026-02566-6}

Wise, S. L., \& Kong, X. (2005). Response time effort: A new measure of examinee motivation in computer-based tests. \emph{Applied Measurement in Education, 18}(2), 163--183. \url{https://doi.org/10.1207/s15324818ame1802_2}

Yin, M., Wortman Vaughan, J., \& Wallach, H. (2019). Understanding the effect of accuracy on trust in machine learning models. In \emph{Proceedings of the 2019 CHI Conference on Human Factors in Computing Systems} (Paper 279). ACM. \url{https://doi.org/10.1145/3290605.3300509}

\par\endgroup

\clearpage\setcounter{table}{0}\renewcommand{\thetable}{S\arabic{table}}

\hypertarget{supplementary-material-s1.-simulation-details}{%
\section{Supplementary Material S1. Simulation details}\label{supplementary-material-s1.-simulation-details}}

Each of the 1,000 parameter sets draws the values in Table S1 independently. Within a parameter set, the same 300 simulated people, with the same self-confidence, the same effort per check, the same wrong verdicts and the same test claims, take part in every condition, so that differences between conditions come from the conditions alone. The design is that of Section 5: 24 practice claims, 8 on each of three rules, interleaved and 120 seconds apart, in three sessions a week apart, and a test 90 days after the last session with 6 new claims per rule, 2 for each warranted response, on the three practised rules and one unpractised rule. With a copyable task, a person checks a claim with probability

\begin{equation}
h_{ik} = \frac{1}{1 + e^{-(\eta_{0} + \eta_{1}(1 - \hat{\rho}_{ik}) + c_{i} - \ell_{ik})}}
\end{equation}

During practice, a person who does not check adopts the verdict, and a person who checks responds as at the test, with an extra pull \(g\) towards the response the verdict gives. The ranges are assumptions, not estimates. They were chosen so that delayed unaided judgment lies between chance and ceiling in every condition (across conditions and parameter sets it ranged from .36 to .88, where chance is .33), so that people told that the assistant is always right still check some claims (from 5\% to 31\%, 16\% on average), and so that the gains are neither negligible nor at their maximum. Across parameter sets, the gains were largest where an explanation read without a check taught little (rank correlations with \(\varepsilon\) of −.44 for the task gain, −.49 for the fallibility gain and −.48 for the belief gain), where forgetting was fast (with \(\alpha\), .43, .37 and .31), where rules were stronger before practice (with the mean of \(\beta\), .39, .30 and .24) and more readily recognised (with the offset of \(\tau\), .35, .28 and .25), and where a recognised rule pulled harder towards the warranted response (with \(\lambda\), about .20). The task gain was also larger where people checked an assistant believed always right less often (with \(\eta_{0}\), −.30), and the fallibility and belief gains rose with \(\eta_{1}\) (.26 and .33) and with the error rate (.19 and .25), whereas \(\psi\) and \(\delta\) hardly mattered (correlations of .04 or less).

\textbf{Sensitivity analysis.} One assumption at a time was varied over the same parameter sets. With \(\varepsilon\) fixed at .3, the medians of the task gain, the fallibility gain and the belief gain were 2.9, 1.7 and 1.1 points; at .6 they were 1.0, 0.5 and 0.3; and at .9 they were 0.2 points or less. With a discount ω on every explanation from the first error a person sees, the fallibility gain with the copyable task fell to medians of 3.3, 2.6 and 1.2 points at ω = .1, .25 and .5, and reversed in 0\%, 2\% and 24\% of parameter sets; with the check-requiring task, the erring assistant taught less than the always-right one by medians of 0.2, 0.5 and 1.4 points, and by more than 1 point in 1\%, 23\% and 63\% of sets. Compliance was not varied, but its effect can be read from the task gain, which amounted to a median of 0.7 points for every 10 percentage points of claims checked (10th to 90th percentile, 0.2 to 1.6); a difference of 10 points in the rate of earnest answering between reliability conditions would thus move prediction 3 by roughly as much as a discount of ω = .25. With \(\kappa\) in the bands 2 to 10, 10 to 20 and 20 to 30 verdicts, the belief gain was .53, .71 and .79 of the fallibility gain, and the share of claims checked with the stated rate fell between the first and the third session from 44\% to 25\%, from 50\% to 32\% and from 52\% to 38\%. Replacing the logistic \(h\) by a linear or a threshold function changed the sizes of the gains but never reversed their signs. When a check that fails, that is, one followed by a wrong judgment during practice, was given the weight .5 instead of 1, the task, fallibility and belief gains were 5.8, 3.5 and 2.3 points, B-X still equalled E-X exactly, and F-X fell below E-X by a median of 0.1 points (10th to 90th percentile, 0.0 to 0.4), because wrong verdicts made checks fail more often.

\textbf{Alternative accounts.} A further script adds the belief-only condition with the check-requiring task (B-X) and simulates two alternative accounts on the same parameter sets. In the depth or feedback-encoding account, the weight of every trace, from a check or from an explanation read without one, is multiplied by 1 + θ(1 − \(\hat{\rho}\))/(1 − \(\rho_{0}\)), so that a person whose perceived reliability equals the stated rate learns 1 + θ times as much from each; θ = 0 is strict VLT. For θ = .1, .25, .5, 1 and 2, B-X exceeded E-X by medians of 0.1, 0.3, 0.5, 0.9 and 1.4 points (10th to 90th percentile at θ = 1, 0.1 to 2.8; above 2 points in 18\% of sets at θ = 1 and 34\% at θ = 2), and F-X exceeded E-X by 0.2, 0.4, 0.7, 1.1 and 1.6 points, while the belief gain with the copyable task rose to 2.7, 3.2, 3.7, 4.5 and 5.6 points; the task-by-belief interaction was positive in every parameter set. With three practice claims per rule instead of eight, B-X exceeded E-X by medians of 1.4 points at θ = .5 and 2.2 points at θ = 1, against belief gains of 7.2 and 8.7 points with the copyable task. With responses sampled, balanced accuracy on the 18 test claims for practised rules had a person-level standard deviation of 11.9 points with the always-right assistant and the copyable task, and 11.2 points with the check-requiring task (medians over parameter sets); an equivalence test of two independent groups with a margin of 1, 2 or 3 points (two one-sided tests at \(\alpha\) = .05, power .80, true difference zero) then needs about 2,100, 530 or 240 people per condition. In the response-tendency account, an erring assistant leaves learning unchanged and lowers every person's tendency to accept at the test by \(x\) on the log-odds scale. With \(x\) = .25, .5 and 1, accuracy on claims that warrant acceptance fell by 4.9, 9.8 and 19.4 points on practised rules and by 4.9, 9.4 and 17.5 points on the unpractised rule; accuracy on claims that warrant rejecting or withholding rose by 2.3, 4.3 and 7.5 points on practised rules and by about 2.4, 4.5 and 8.2 points on the unpractised rule; and balanced accuracy fell slightly (medians of −0.1, −0.5 and −1.7 points on practised rules), because the simulated people had no average lean towards any response. The script also confirms that strict VLT gives differences of exactly zero for B-X − E-X and F-X − E-X, and reports the gains for tests 7, 30, 90, 180 and 365 days after practice: the task gain rose from a median of 4.3 points at 7 days to 5.8 at 90 and 6.1 at 365 days, the fallibility gain from 2.8 to 3.6 and 3.9, and the belief gain from 1.9 to 2.4 and 2.6.

\textbf{Extra checks.} Because the decay rate of a trace rises with the rule's strength, an extra check also raises the decay rate of the traces laid down after it, so it need not raise the rule's later strength. With \(\delta\) = 2 and \(\alpha\) = .1, far outside the simulated ranges, two checks an hour apart leave a rule weaker an hour after the second than the second check alone would (strength −1.48 against −1.41, with \(\beta\) = −3). Within the simulated ranges, in a grid of 124,416 combinations of parameter values and schedules, an extra early check lowered later strength in 38 cases, only when a run of 15 or more checks followed it one second apart, and then by less than \(10^{-7}\). In the 1,000 parameter sets of the simulation, an extra check on the first practice claim of a rule never lowered the rule's strength at 90 days.

\begin{table}[tbp]
\caption{Parameters of the simulation and the values drawn}
\small\setstretch{1.0}
\begin{tabular}{@{}>{\raggedright\arraybackslash}p{0.085\linewidth}@{\hspace{1.4em}}>{\raggedright\arraybackslash}p{0.565\linewidth}@{\hspace{1.4em}}>{\raggedright\arraybackslash}p{0.275\linewidth}@{}}
\toprule
Symbol & Meaning & Values \\
\midrule
1 − \(\rho_{0}\) & Stated error rate; when the assistant errs, the actual rate rounds it to a whole number of errors among the 24 practice claims & Uniform, .10 to .30 \\ \addlinespace[3pt]
\(\kappa\) & Weight of the statement, in verdicts, the same for everyone within a parameter set & Uniform, 2 to 30 \\ \addlinespace[3pt]
\(\eta_{0}\) & Log-odds of a check with an assistant believed always right, for an average person and claim & Uniform, −3 to −1 \\ \addlinespace[3pt]
\(\eta_{1}\) & Rise in that log-odds per unit of perceived unreliability, 1 − \(\hat{\rho}\) & Uniform, 5 to 20 \\ \addlinespace[3pt]
\(c\) & Confidence in one's own judgment, for each person & Normal, mean 0, SD uniform from .3 to 1 \\ \addlinespace[3pt]
\(\ell\) & Effort a check takes, for each person and claim & Normal, mean 0, SD .5 \\ \addlinespace[3pt]
\(\varepsilon\) & Weight of an explanation read without a check & Uniform, 0 to .3 \\ \addlinespace[3pt]
\(\delta\) & Decay scale & Uniform, .15 to .30 \\ \addlinespace[3pt]
\(\alpha\) & Decay intercept & Uniform, .10 to .25 \\ \addlinespace[3pt]
\(\psi\) & Weight of time between sessions & Uniform, .01 to .10 \\ \addlinespace[3pt]
\(\beta\) & Strength of a rule before practice, for each person and rule & Normal, mean uniform from −4 to −2.5, SD .5 \\ \addlinespace[3pt]
\(\tau\) & Recognition threshold & Mean of \(\beta\) plus a value uniform from .5 to 1.5 \\ \addlinespace[3pt]
\(\sigma\) & Noise in strength & Uniform, .3 to .7 \\ \addlinespace[3pt]
\(\lambda\) & Pull of a recognised rule towards the warranted response & Uniform, 1 to 3 \\ \addlinespace[3pt]
\(b\) & Tendency towards each response, for each person & Normal, mean 0, SD .4 \\ \addlinespace[3pt]
\(\gamma\) & Pull of each claim towards each response, for each person & Normal, mean 0, SD uniform from .3 to 1 \\ \addlinespace[3pt]
\(g\) & Pull of the verdict on a person who checks, during practice & Uniform, 1 to 3 \\
\bottomrule
\end{tabular}
\end{table}

\hypertarget{supplementary-material-s2.-a-proposed-measurement-model-for-checks}{%
\section{Supplementary Material S2. A proposed measurement model for checks}\label{supplementary-material-s2.-a-proposed-measurement-model-for-checks}}

This supplement proposes a measurement model for the checks on which prediction 7 depends. It has not been validated; Section 6 states what validation requires. Prediction 7 needs to know, for each person and practice claim, whether a check was made. The measurement model treats this as a latent variable \(C\), equal to 1 for a check and 0 otherwise, and links it to two kinds of indicators. The prior probability of a check is estimated for each condition, since VLT expects it to differ between them. The inferred checks then carry the condition's information only through how often people check, which is how VLT says the condition acts, and the incremental test of prediction 7 is informative only if the indicators make the inferred checks vary within conditions.

\emph{Shared indicators} are recorded in the same way in every condition, which makes them the basis for comparing conditions: whether the person opened or dwelt on the part of the evidence that the rule concerns before the explanation appeared, and the time spent on the evidence after the verdict. The first requires the evidence to be shown in parts, which must be the same and unlabelled in every condition, since labels naming what each part concerns would draw attention to the rules' conditions (Section 6); showing the evidence in parts may itself raise checks in every condition. Time is modelled as a mixture of two distributions, one for claims with a check and one for claims without; very short times mark responses made without examining the evidence, as response times mark rapid guessing in tests (Wise \& Kong, 2005). The measurement parameters of the shared indicators must be the same in every condition compared, which a test should check rather than assume. Because a flip question makes people dwell on the evidence whether or not they check, this is plausible only within one task, which is a further reason to run the incremental test of prediction 7 within one task.

\emph{Task-specific indicators} are available only in some conditions. Under a check-requiring task with the fixed options of Table 1, each flip answer is coded in one of three ways: as an attempt at the rule the claim calls for, if it changes the feature that rule concerns, whether or not the change alters the warranted response; as an attempt at another rule, if it changes a feature another rule concerns, which can be credited as a trace on that rule but is not a check of the rule the claim calls for; or as a guess, if it is given faster than the evidence can be read. The rate at which guesses happen to be correct follows from the number of options. An attempt that reaches the wrong answer is a check that failed; the strict model counts it in full, a simplification discussed in Section 3. With a copyable task, an override of a wrong verdict is evidence of a check, but noisy evidence: a person who distrusts the assistant can override it without checking and, with three responses, will then choose the warranted response half the time. Overrides of wrong verdicts should therefore be read against how often the same person overrides right ones, and since they can occur only when the assistant errs, they should never be the only indicator in a comparison.

The graded form of the model (Section 3) adds the depth \(D\) of a check, for claims on which \(C\) = 1, measured by process indicators alone, such as the time spent on the part of the evidence the rule concerns, and not by whether the check reached the right answer, since a check counts whether or not it succeeds. Indicators of depth other than time, such as how much of the evidence a flip answer takes into account, exist only under the flip task, which limits comparisons of depth between tasks. A superficial check is then one of low depth, which the strict model counts in full and the graded form counts in part.

The measurement model and the memory model are estimated together, so that uncertainty about the checks carries into the forecasts. For each draw of the latent checks from their posterior distribution, the memory model computes the strength of each rule at the test, and the forecasts average over the draws. For the people whose data fix the parameters, the checks are inferred from all their data; for the people on whom the forecasts are scored, they are inferred from the practice data alone, without the test responses. The incremental test of prediction 7 compares forecasts with and without the condition across these draws, by the scoring rule fixed in advance. Before any data are collected, simulating data from the model with known parameters and estimating them again shows whether the parameters can be recovered with the planned numbers of people and claims. Posterior predictive checks of the indicators in each condition then show whether the measurement model fits equally well in all of them; a misfit confined to one condition would undermine the comparison.

\end{document}